\documentclass{article}

\usepackage[preprint]{neurips_2020}

\usepackage[utf8]{inputenc} 
\usepackage[T1]{fontenc}    
\usepackage{url}            
\usepackage{booktabs}       
\usepackage{amsfonts}       
\usepackage{nicefrac}       
\usepackage{microtype}      
\usepackage{amsmath}
\usepackage{amssymb}
\usepackage{bm}

\usepackage{amsmath,amsfonts,bm}

\def\eqref#1{equation~\ref{#1}}

\def\1{\bm{1}}

\def\vtheta{{\bm{\theta}}}

\def\vv{{\bm{v}}}

\def\vx{{\bm{x}}}
\def\vy{{\bm{y}}}

\DeclareMathAlphabet{\mathsfit}{\encodingdefault}{\sfdefault}{m}{sl}
\SetMathAlphabet{\mathsfit}{bold}{\encodingdefault}{\sfdefault}{bx}{n}

\def\sX{{\mathbb{X}}}

\DeclareMathOperator*{\argmin}{arg\,min}

\usepackage{xspace}

\makeatletter
\DeclareRobustCommand\onedot{\futurelet\@let@token\@onedot}
\def\@onedot{\ifx\@let@token.\else.\null\fi\xspace}
\def\eg{\emph{e.g}\onedot} \def\Eg{\emph{E.g}\onedot}
\def\ie{\emph{i.e}\onedot}

\makeatother

\usepackage{mathtools}
\usepackage{multirow}
\usepackage{caption}

\usepackage{times}
\usepackage{epsfig}
\usepackage{graphicx}
\usepackage{subfigure}
\usepackage{xcolor}
\usepackage{bbding}
\usepackage{pifont}
\usepackage{amsthm}
\usepackage{enumitem}

\definecolor{citecolor}{rgb}{0.133, 0.752, 0.133}
\usepackage{makecell}
\usepackage[pagebackref=true,breaklinks=true,letterpaper=true,colorlinks,bookmarks=false,citecolor=blue,linkcolor=blue]{hyperref}
\definecolor{Highlight}{HTML}{39b54a}  

\usepackage{algorithm}
\usepackage{listings}
\usepackage{pdfpages}
\usepackage{etoolbox}
\makeatletter
\AfterEndEnvironment{algorithm}{\let\@algcomment\relax}
\AtEndEnvironment{algorithm}{\kern2pt\hrule\relax\vskip3pt\@algcomment}
\let\@algcomment\relax
\newcommand\algcomment[1]{\def\@algcomment{\footnotesize#1}}
\renewcommand\fs@ruled{\def\@fs@cfont{\bfseries}\let\@fs@capt\floatc@ruled
  \def\@fs@pre{\hrule height.8pt depth0pt \kern2pt}%
  \def\@fs@post{}%
  \def\@fs@mid{\kern2pt\hrule\kern2pt}%
  \let\@fs@iftopcapt\iftrue}
\makeatother

\newlength\savewidth\newcommand\shline{\noalign{\global\savewidth\arrayrulewidth
  \global\arrayrulewidth 1pt}\hline\noalign{\global\arrayrulewidth\savewidth}}
\newcommand{\tablestyle}[2]{\setlength{\tabcolsep}{#1}\renewcommand{\arraystretch}{#2}\centering\footnotesize}

\def\degree{${}^{\circ}$}

\title{IDEAL: Independent Domain Embedding Augmentation Learning}

\author{
	Zhiyuan Chen,  \
	Guang Yao,  \
	Wennan Ma, \
	Lin Xu\thanks{Contact Author (Email: lin.xu5470@gmail.com)} \\
	Shanghai Em-Data Technology Co., Ltd. \\
}

\begin{document}
	
	\maketitle	
	\begin{abstract}
    Many efforts have been devoted to designing sampling, mining, and weighting strategies in high-level deep metric learning (DML) loss objectives. However, little attention has been paid to low-level but essential data transformation. In this paper, we develop a novel mechanism, independent domain embedding augmentation learning ({IDEAL}) method. It can simultaneously learn multiple independent embedding spaces for multiple domains generated by predefined data transformations. Our IDEAL is orthogonal to existing DML techniques and can be seamlessly combined with prior DML approaches for enhanced performance. Empirical results on visual retrieval tasks demonstrate the superiority of the proposed method. For example, the IDEAL improves the performance of MS loss by a large margin, 84.5\% $\rightarrow$ 87.1\% on Cars-196, and  65.8\% $\rightarrow$  69.5\% on CUB-200 at Recall$@1$. Our IDEAL with MS loss also achieves the new state-of-the-art performance on three image retrieval benchmarks, \ie, \emph{Cars-196},  \emph{CUB-200}, and \emph{SOP}. It outperforms the most recent DML approaches, such as Circle loss and XBM, significantly. The source code and pre-trained models of our method will be available at
	\emph{\url{https://github.com/emdata-ailab/IDEAL}}.
	\end{abstract}
	
\section{Introduction}
Metric learning aims to learn a semantic embedding space, where similar samples are encouraged to be close together, while dissimilar ones are far apart. With the flourish of deep learning, deep metric learning (DML) has drawn much interest in recent years \cite{contrastive,Hoffer2015DeepML,schroff2015facenet, lifted-structured-loss}. 
Exploring and collecting informative negatives is of great importance in DML. To reach this, many efforts have been devoted to designing complicated pairwise sampling, mining or weighting strategies to capture more information during training, \eg, distance-weighted sampling in \cite{wu2017sampling}, log-sum-exp weighting schemes in \cite{lifted-structured-loss, n-pairs}, and constructing pairs across preceding mini-batches from cached embeddings \cite{wang2019cross}. 
  
Despite the significant progress, existing works in DML often focus on designing high-level learning objective functions, while ignoring the low-level but essential \emph{data transformation}. In fact, data transformation has played a crucial role in recently advanced methods in self-supervised representation learning. In contrastive learning, SimCLR \cite{simclr} experimentally demonstrated that the composition of multiple data transformations is of central importance in learning effective representations. Moreover, in pretext learning, it yields a series of pretext tasks to predict what the specific transformation is applied to an image, \eg, rotation estimation \cite{gidaris2018unsupervised} and patch orderings prediction \cite{noroozi2016unsupervised}. 

In reality, the requirements for applying data transformations are not always satisfied. For pretext prediction task, the basic premise is that the original data is \emph{transformation determinable}. \Eg, rotation prediction can be applied in landscape datasets, but not medical images of cells, which is not rotation determinable. For data augmentation, another use of data transformations, expert knowledge needs. For example, random rotation may improve the robustness of the model on medical cell images, while on \emph{MNIST}, it can cause serious problems, \eg, semantic confusion in numbers $6$ and $9$.  

In this paper, to exploit the power of data transformations in DML, we develop a novel method that can apply data transformations without such limitations as in pretext prediction or data augmentation. 
First, we use \emph{Domain Augmentation} to generate multiple domains with a set of predefined transformations. Each domain can be viewed as a new view of the dataset. From this perspective, a naive solution, following conventional multi-view learning, is to train different models for each domain. This solution is compute-intensive and time-consuming. Inspired by multi-task learning using different branches to optimize different tasks, we propose our Independent Domain Embedding Augmentation Learning (IDEAL) to train one model by optimizing the embedding spaces on all the domains simultaneously. Our IDEAL can be regarded as sharing parameters between multiple independent models. Such parameter sharing benefits the generality and robustness of the trained model, which is verified in our experiments. The main contributions of this paper are summarized as follows:

{
\setlength\parindent{1em} $ \bullet$   We propose IDEAL, a simple yet general approach for DML, that leverages \emph{Domain Augmentation} to jointly learn a more robust model with multi-tasks learning fashion. 

\setlength\parindent{1em} $ \bullet$  We discuss the derived properties of our method and compare it with existing general mechanisms to conventional practices that demonstrate the superiority of our scheme theoretically and empirically.

\setlength\parindent{1em} $ \bullet$  We evaluate our IDEAL with various conventional DML losses on  widely used image retrieval datasets: \emph{Cars-196}, \emph{CUB-200}, and \emph{SOP}, surpassing the state-of-the-art methods by a large margin.	
}

	\section{Related Work}
Our work draws on existing literature in contrastive learning, ensemble learning, and multi-task learning. Due to a large amount of literature, we focus on the most relevant papers.
	
	\textbf {Contrastive Learning.} 
    As the fundamental method in metric learning, contrastive learning aims to learn an
embedding space where similar instances can be distinguished from dissimilar data, and are also at the core of several recent works on self-supervised representation learning \cite{wu2018unsupervised,henaff2019data,hjelm2018learning,tian2019contrastive,sermanet2018time,simclr}.
Then according to computing similarity metric on real samples or virtual proxies, these methods could be divided into pair-based \cite{lifted-structured-loss, wang2019multi, contrastive} or proxy-based \cite{movshovitz2017no, sun2020circle, wang2019cross, qian2019softtriple} branches, respectively. 
In the contrastive learning setting, positive and negative pairs are generated in mini-batch of samples for each anchor image, often using hard mining. Different contrastive learning methods usually vary in their specific sampling and weighting strategy to design loss function, such as distance-weighted sampling in \cite{wu2017sampling}, sampling informative instances from memory bank in \cite{wang2019cross}, uniformly weighting in vanilla contrastive loss \cite{contrastive}, log-sum-exp weighting in N-pair loss \cite{n-pairs} and InfoNCE loss\cite{he_2019_moco}.

	\textbf {Ensemble Learning.} 
    Another slightly related to our approach in metric learning is the ensemble learning \cite{yuan2017hard,opitz2018deep,li2020divide}.  HDC \cite{yuan2017hard} ensembles a set of models with different complexities in a cascaded manner for mining hard examples at multiple levels. A-BIER \cite{opitz2018deep} divides the last embedding layer of a deep network into an embedding ensemble and formulate training this ensemble as an online gradient boosting problem. \cite{li2020divide} follows a divide-and-conquer strategy by splitting and subsequently merging both the data and embedding space. 
    The critical difference in our IDEAL is that we divide embedding subspace based on corresponding augmented domains. The model is independently trained on non-overlapping augmented domains in multi-task learning fashion. Finally, to fully exploit the value of each domain, we adopt an ensemble strategy during testing.
	
	\textbf {Multi-task Learning.} 
	Multi-task learning has emerged as a promising approach for sharing structure across multiple tasks to enable more efficient learning. Multi-task deep metric learning (MDML) has also recently been shown to be advantageous, allowing to learn the metrics for several related tasks jointly \cite{parameswaran2010large, milbich2020diva}. In MDML, concurrently solving different tasks is also employed by classical multi-task learning and often based on a divide-and-conquer principle with multiple learners optimizing a given subtask \cite{milbich2020diva}. For example, \cite{bhattarai2016cp} utilizes additional training data and annotations to capture extra information, while our IDEAL is defined on standard training data only. In practice, multi-task learning presents a challenging optimization problem, which has been tackled in several ways in prior work. A number of architectural solutions have been proposed to the multi-task learning problem based on multiple modules or paths \cite{rosenbaum2017routing}. It inspired us to formulate various augmented domains to create multiple auxiliary learning tasks, and employ an independent branch for each domain. 
	


	\section{Preliminary} 
When training with data augmentation, the model is encouraged to learn a representation invariant to data augmentation. Therefore, data transformations, \eg, random cropping, image mirroring, and color-shifting, are applied under the requirement of preserving the images' original semantics. As these transformations designed manually, it requires expert knowledge and sufficient experiments. For example, horizontal flipping of images during training is a useful data augmentation method on \emph{CIFAR-10} but not on \emph{MNIST}, due to the different symmetry properties implicate in datasets.  
In other words, for a specific dataset, where the manifold of images (i.e., source domain) has its own specific transformation invariant property. Under these specialized transformations, the manifold of images would keep unchanged. Only these transformations could be used as effective data augmentation.

Horizontal flipping and random cropping are widely used data augmentation in the \emph{ImageNet} \cite{ILSVRC15}. Because the mirror symmetry is intrinsic property among natural objects, the manifold of \emph{ImageNet} does not have an essential difference with the source domain under horizontal flipping transformation. Similarly, random cropping and resizing can be considered as adjusting the distance and position of the object from the photographer, thus it could not change the manifold of \emph{ImageNet} as well. However, horizontally flipping is useless and even harmful data augmentation for \emph{ImageNet} because of no vertical symmetry among most natural objects. Most images are `up-standing' as this is how humans look at them, which is implicit bias introduced from the photographer. Vertically flipping the source domain of ImageNet yields an entirely different data domain, a `down-standing' domain. We defined such augmentation as Domain Augmentaion. Hence, for a specific dataset, the two sets of transformations used in data augmentation and domain augmentation are mutually-exclusive. Disregarding such differences of domains during training would degrade convergence.

Another line of augmentation methods focuses on generating a series of prediction pretext tasks that form pseudo-labels by specific transformation, such as prediction of orderings \cite{doersch2015unsupervised,noroozi2016unsupervised} and transformations estimation \cite{dosovitskiy2014discriminative,gidaris2018unsupervised,caron2019unsupervised,zhang2019aet}. In contrast with data augmentation, all these transformations break the invariant property, \ie, domain augmentation is exactly an adequate transformation in pretext task. Thus, both vertically flipping and rotation could be used in transformations prediction tasks for ImageNet \cite{gidaris2018unsupervised}, because vertically flipping and rotation are \textbf{domain augmentation} for ImageNet by changing the canonical `up-standing' orientation. Nevertheless, the rotation transformations are \textbf{data augmentation} for some orientation agnostic datasets, \eg, medical cell images or satellite images. Regarding this cases, the pseudo-labels provided by such transformation can be seen as noisy labels. The pretext task of classifying such noisy labels is ill-posed and would degrade performance.

From the above limitations, it inspired us to leverage domain augmentation in a new training fashion, which is distinctive from the training mechanism in the pretext task and data augmentation methods. Intuitively, since domain augmentation brings different data domains, we attempt to avoid learning an invariant representation as conventional data augmentation fashion. We note that each domain has its own semantic space. In the case of \emph{MINST}, the semantic spaces between the source domain and the vertically flipping domain are entirely different. The number of $6$ in the source domain has the same semantic as the number of $9$ in the flipping domain. Therefore, we believe that maintaining the independent property of semantic embedding spaces between these domains is the key for training. 
In the next section, we shall develop our IDEAL method motivated by this observation.

	\section{Our Method}
	
	\subsection{Independent Domain Embedding Learning}
	Let $\sX=\{\vx_1, \vx_2, \dots, \vx_N\}$ denotes the training set, and $y_i$ be the corresponding label of $\vx_i$. An embedding function, $f(\cdot; \vtheta)$, to project a data point $\vx_i$ onto a $d$-dimensional space, $\vv_i = f(\vx_i; \vtheta)$, where $\vtheta$ is the parameters of $f$. The similarity between two instances are measured by the cosine similarity in the embedding space. 
	
	Let $T$ be a domain augmentation, such as image rotation of 90 degrees, then $T(\sX)=\{ \ T(\vx_1), T(\vx_2), \dots, T(\vx_N) \ \}$ is the augmented domain of $\sX$ by $T$. We use $\mathcal{T} = \{T_{0}, T_{1}, \ldots, T_{k-1} \}$ to represent a set of domain augmentations, then applying $T_i$ on $\sX$ can generate k \emph{independent} domains: $\{\ T_{0}(\sX), T_{1}(\sX), \ldots, T_{k-1}(\sX) \ \}$. To be concise, we set $T_0$ to be the identity transform as default, thus $T_{0}(\sX) = \sX$, is the original domain.

	In fact, the augmented domain can be seen as a new view of data. Thus, An evident way to leverage such augmentation is to train multiple independent models for each view as conventional multi-view learning schemes. However, this solution is obviously compute-intensive and time-consuming. We seek an alternative: training one model by optimizing the embedding spaces on all the domains simultaneously. Formally, we train the model by optimizing the multiple objectives as below:
	
	\begin{equation}
	\argmin_{\vtheta} \mathcal{J}(f(\sX_i, \vtheta), \vy), \quad for \  i \in \{0, 1, \ldots, k-1\},
	\end{equation}
	where the concrete formulation of objectives $\mathcal{J}$ can be derived from any existing DML method, \eg, contrastive loss \cite{contrastive}, triplet loss \cite{schroff2015facenet}, and MS loss \cite{wang2019multi}. For example, if using triplet loss during training, we uniformly sample a mini-batch $B$ and follow our learning fashion as:
	
	\begin{equation}
	\mathcal{L}(B)=\sum\limits_{T \in \mathcal{T}_{i}} \sum_{\substack{(a, p, n) \\ \sim {T} \circ B}}\left[ \left\|f_{a}-f_{p}\right\|_{2}-\left\|f_{a}-f_{n}\right\|_{2}+\alpha \right]_{+},
	\end{equation}
	
	where $[\cdot]_+$ denotes the positive part, and $\alpha$ is the margin. We use $(a, p, n) \in {T} \circ B$ as the triplets sampled from the current mini-batch transformed with given domain augmentation ${T}$.  The triplet loss strives to keep the anchor point $a$ closer to the positive data point $p$ than the negative point $n$ by at least a margin $\alpha$. For brevity, we omit the definitions of contrastive loss and MS loss, but we refer the interested reader to the original works \cite{contrastive, wang2019multi}. 
	
	Intuitively, our method described above serves to generate one surrogate task for each data domain. At the same time, it enforces different data domains to learn an exclusive embedding space with their respective losses independently. In our work, for image retrieval datasets, we adopt the set of Domain Augmentation  $\mathcal{T}$ as image rotations by multiples of $90$ degrees, \ie, $\mathcal{T}=\{ T_{0}, \ldots, T_{3} \}$. More formally, if $\operatorname{Rot}(x, \phi)$ denotes rotates image $x$ by $\phi$ degrees, then $T_{i}=\operatorname{Rot}(x, 90 \cdot i)$.


	
	\textbf{Inference.} We expected that multiple domains could provide complementary information. To fully exploit the value of each domain, we attempt to adopt an ensemble strategy during testing. Given a query or gallery image $\vx$, we concatenate the features extracted from its corresponding instances in different domains and measure the similarity of ensemble features by dot product. This ensemble strategy is equivalent to calculating respective similarities in each domain, and then retrieving based on their aggregated similarity.

	In the following sections, we provide a foundation for this intuition. We then discuss the derived properties of this approach and compare it to standard practices for metric learning.

	\begin{figure*}[htb]
		\centering
		\includegraphics[width=1\linewidth]{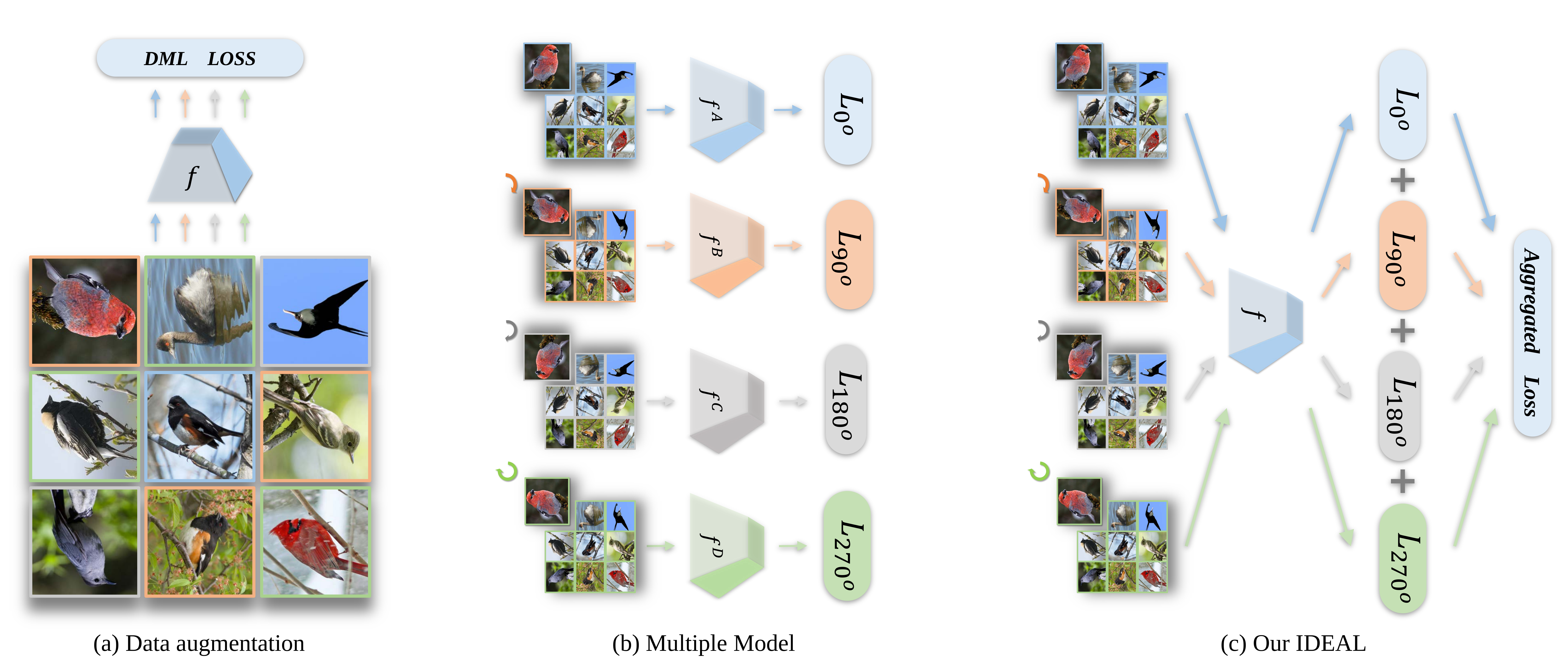}
		\caption{Conceptual comparison of three previous mechanisms (empirical comparisons are in Tab.\ref{baseline}). Colors (\ie, {\color[rgb]{0.4, 0.65, 0.86}  blue}, {\color[rgb]{0.94, 0.57, 0.32} orange}, {\color[rgb]{0.6, 0.6, 0.6} gray} and {\color[rgb]{0.59, 0.78, 0.48} green}) represent different domains (\ie, $0$, $90$, $180$, $270$ degrees generated from corresponding rotation transformation). \textbf{(a) Data Augmentation Mechanism:} The model is encouraged to learn a representation invariant to data augmentation. \textbf{(b) Multiple Model Mechanism:} Optimizing multiple tasks of different domains individually, 
		each model is trained to learn an exclusive embedding space independently. \textbf{(c) The proposed IDEAL:} Our IDEAL can learn a single parameter-sharing model among multiple independent domains. The embedding spaces on all the  domains can be optimized simultaneously.}
		\label{fig:comparison}
		\vspace{-0.3cm}
	\end{figure*}

	\subsection{Conceptual Comparison to Previous Training Mechanisms} 	
	\textbf{Data Augmentation Training Mechanism.} 
	In the training procedure with data augmentation, the goal is to learn a representation $\Phi$ which is invariant to transformation $T$, \ie, $\Phi(x) = \Phi(T(x))$ as Fig.\ref{fig:comparison}a. Obviously, training with domain augmentation in this manner may deteriorate performance, because the relations among different domains get entangled. In some cases, the discrepancy of domains is extreme: pulling the positive pair with identical class labels from different domains to the same semantic embedding is meaningless and even destructive. Similarly, it is also questionable whether two pictures augmented from different categories can be taken as negative pair naively (such as the specific scenarios of the number of $6$ and $9$). 
	
	\textbf{Multiple Model Training Mechanism.}
	To circumvent the above issue, we could learn an independent model for each domain, and each model is trained to learn an individual embedding space independent from the others as Fig.\ref{fig:comparison}b. Though optimizing each model individually can mitigate detrimental gradient interference between different domains, it is obviously computationally intensive and impractical. In our IDEAL, for computational efficiency, we learn a single model by sharing the parameters among multiple domains. Moreover, our experimental results verify such parameter sharing could facilitate the generality and robustness of the trained model.
	
	
	


	\textbf{Pretext Augmentation Training Mechanism.}
    The pretext task is an unsupervised method, but our IDEAL is a general learning fashion that could be integrated into a supervised or unsupervised deep metric learning task flexibly. Moreover, our IDEAL breaks the limitation of pretext task, \ie, the basic premise of rotation prediction is that the natural images must have a canonical image orientation, and the neural network has to recognize and localize the object in the image. However, the prerequisite introduced in these methods could not be satisfying for some natural images, where the image orientation is ambiguous, \eg, ball and cell images. 
	In our approach, the prerequisite of previous works become unnecessary, because we do not resort to the discrimination of specific transformations of an image. Empirically, the advantages of our method can also be demonstrated in rotation agnostic datasets (\eg, SOP dataset, many images are the local parts of products such as bicycle wheel). We argue that more relevant domains augmented from rotation insensitive data benefits embedding learning because of more consistent gradient aggregated from multiple domains.
	
	\subsection{Independence Exploitation}
    Compared to data augmentation training mechanism, our IDEAL only achieved the disentanglement of domains at the level of the loss function, \ie, to cancel the harmful gradients arising from those positive and negative pairs constructed from different domains. However, multiple model training mechanism not only reached this goal, but also achieved the disentanglement at the level of model structure. Thus, the independence property among domains may be damaged by sharing parameters. 
    
    To overcome this problem, inspired by multi-task learning using multiple branches to optimize different tasks, we employ an independent branch for each domain. To avoid introduce any additional parameters, we divide the last embedding layer of the model into four non-overlapping heads as in BIER \cite{opitz2018deep}. In this manner, we could assign one independent embedding subspace to each domain. Specifically, each augmented image can only forward through their corresponding head, and each head's parameters would only be updated by the domain it is responsible to. Empirically, our IDEAL leads to additional gains in performance with this independence design at model structural level.

    In fact, such a design can be considered a relaxant of sharing the whole model's parameters. Note that the first few layers of most networks extract low-level information, which is invariant to our domain augmentation. On the other hand, we conjecture that inconsistent gradient aggregated from different domains more likely to appear in high-level layers. Thus, sharing parameters only in the low-level layers is a better choice, maintaining the independent property among domains to a deeper degree.

	\section{Experiments}
	\label{others}
	We  describe the training details and the benchmark datasets. We then performed extensive experiments to validate the derived properties of our method. Finally, we evaluate our IDEAL with various conventional DML losses on standardized image retrieval datasets.

	\subsection{Implementation Details}
	
	For a fair comparison to previous work, the Inception network with batch normalization \cite{batchnorm} pre-trained on \emph{ILSVRC2020-CLS} \cite{ILSVRC15} is adopted as our backbone. We add a 512-d fully-connected layer as a projection head following the global pooling layer and L2-normalize the final output. All input images are first resized to $256 \times 256$, and then cropped to $224 \times 224$. Random crops and random flips are utilized as data augmentation during training. The test set was preprocessed just by the center crop. For all experiments, we use Adam optimizer with $5 e^{-4}$ weight decay and the PK sampler ($8$ categories, $4$ samples/category) during training. The implementation is done using the PyTorch framework, and experiments are performed on compute clusters containing P100 and V100. We utilized the default hyperparameters for contrastive loss, triplet loss and MS loss.

	\subsection{Datasets}
	Our methods are evaluated on three common datasets for few-shot image retrieval (Recall@k is reported). The training and testing protocol follow the standard setups:
	
	\emph{Cars-196 (Cars)} \cite{cars-196} provides $16,185$ images in $196$ categories. The first $98$ classes containing $8054$ images are used for training, while the rest $98$ classes with $8131$ images are used for testing.
	
	\emph{CUB-200-2011 (CUB)} \cite{cub-200}  with $11,788$ bird images from $200$ classes. Following \cite{lifted-structured-loss}, we use $5,864$ images of $100$ classes for training and the remaining $5,924$ images for testing.
	
	\emph{Stanford Online Products (SOP)} \cite{lifted-structured-loss}  provides $120,053$  product images in $22,634$ categories. Following \cite{lifted-structured-loss}, we use $59,551$ images for training, and the remaining $60502$ images for testing.

	\subsection{Detailed Analysis}
	In this section, we performed a series of ablation experiments to validate the derived properties of IDEAL and study the effect of several designed choices in our method. Specifically, we choose the default \textbf{MS-loss} with the \textbf{BN-Inception} network if not mentioned.

	\begin{table}[h]	
		\centering
		\scriptsize
		\setlength\tabcolsep{5.1pt}
		\begin{tabular}{cl|ccccc|ccccc}
			\toprule
			\multicolumn{2}{c}{} & \multicolumn{5}{c}{\textbf{Cars - 196}} & \multicolumn{5}{c}{\textbf{CUB - 200}}\\
			\midrule
			&  Domains       & 0\degree  & 90\degree  & 180\degree & 270\degree & ensemble & 0\degree & 90\degree  & 180\degree & 270\degree  & ensemble \\
			
			\midrule
			\parbox[t]{1mm}{\multirow{4}{*}{\rotatebox[origin=c]{90}{Methods}}}
			& MS loss  & 83.1\% & 45.4\% & 54.5\% & 45.0\% & 80.5\% & 65.1\% & 52.7\% & 51.3\% & 53.6\% & 66.3\%\\
			&Data Augmentation & 78.5\% & 76.5\% & 77.8\% & 76.1\% & 84.3\% &  62.4 \%  &  59.1 \%   &  59.4 \%  & 57.9 \% & 66.8 \% \\
			
			& Multiple Model  & {82.8\%} & {79.2\%} & {82.4\%} & {80.5\%} & 
			\textbf{89.9\%} & 64.3 \%  & 60.1 \% & 60.7 \% & 61.6 \% & \textbf{70.9\%}  \\
		
			& MS w/ IDEAL        & \textbf{84.0\%} & \textbf{80.2\%} & \textbf{82.8\%} & \textbf{80.5\%} & {87.1\%} & \textbf{65.9\%} & \textbf{62.2\%} & \textbf{62.9\%} & \textbf{62.1\%} & 69.5\%  \\
			\bottomrule
		\end{tabular}
		\vspace{0.3cm}
		\caption{Recall$@1$ of several methods on Cars-196 and CUB-200 datasets.}
		\label{baseline}
	\end{table}
	
	\begin{figure}[htb] 
	\centering
	\includegraphics[width=\linewidth]{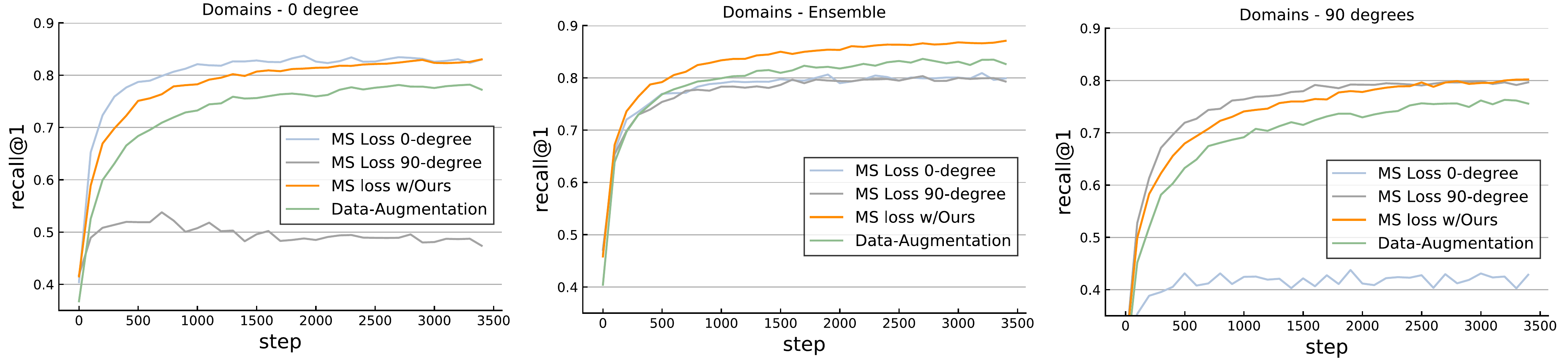}
	\caption{\small{The Recall$@1$ curve of several methods on \textbf{Cars-196}. \textbf{Left}: Several methods evaluated on source domain. \textbf{Middle}: With ensemble strategy. \textbf{Right}: Evaluated on the domain of 90 degrees.}} 
	\label{domainsV1}
\end{figure}
	\subsubsection{Comparison to Data Augmentation Mechanism} 
	We compare the traditional data augmentation mechanism and ours as Fig. \ref{domainsV1}. For this purpose, we evaluate the performance on the test set from different domains (\ie, the test sets of $0$, $90$, $180$, $270$ degrees generated from corresponding rotation augmentation). Tab. \ref{baseline} shows a dramatic drop in performance and convergence from 83.1\% to 78.5\% over baseline in the source (0 degree) test set with data augmentation mechanism, proving forcing representation to be invariant to domain augmentation is detrimental to learning. As illustrated in Fig. \ref{domainsV1}, the data augmentation mechanism shows consistent poor performance among different domains; we attribute this to the embedding space of different domains getting entangled, thus cause adverse optimization.
	Surprisingly, our IDEAL successfully relieved this issue, surpassing the previous mechanism \textbf{with an over 5\% improvement across all domains on Cars-196}, revealing our IDEAL could disentangle the different embedding space from each other.  Furthermore, our IDEAL remarkably boosts the performance of MS loss by 1\% improvement on the source domain and over 30\% on other domains. With our ensemble strategy, our IDEAL could achieve a further 4\% performance gain over MS loss (\ie, \textbf{83.1\% $\rightarrow$ 87.2\%}), but MS loss alone even has a 3\% \textbf{drop} with same ensemble strategy (\ie, 83.1\% $\rightarrow$ 80.5\%).

	\subsubsection{Comparison to Multiple Model Mechanism} 
	Next, we compare the multiple model mechanism and ours with the same BN-Inception backbone and standard embedding dimensionality of $512$. Fig. \ref{domainsV1} shows that each individual model performs well in its responsible domain, which supports our motivation of learning independent embedding space for each domain. The key difference from our IDEAL is lying in sharing parameters or not, which can be regarded as sharing parameters between multiple independent models. However, the multiple model mechanism performs slightly worse in every domain compared to our IDEAL as shown in Tab. \ref{baseline}. 
    This is inline with our hypothesis: aggregating learning signals to optimizing a joint and single model leads to a strong generalization and robustness of our model.
    Moreover, sharing parameters can also be beneficial in reducing extra storage and computation consumption. Furthermore, Fig. \ref{domainsV1} reveals that joint optimization of diverse domains regularizes training, and reduces overfitting effect that often occurs during later stages of training. 
    It is also worth noting that the performance in source ($0$-degree) domain is steady better than other domains, which is probably due to the fact that the pretrained model trained on ILSVRC2020-CLS of 0-degree.

	\subsubsection{Performance Study of Splitting Projection Head}

	\begin{table}[h]	
		\centering
		\scriptsize
		\setlength\tabcolsep{4.7pt}
		\begin{tabular}{cl|ccccc|ccccc}
			\toprule
			\multicolumn{2}{c}{} & \multicolumn{5}{c}{\textbf{Cars - 196}} & \multicolumn{5}{c}{\textbf{CUB - 200}}\\
			\midrule
			&  Domains       & 0\degree  & 90\degree  & 180\degree & 270\degree & ensemble & 0\degree & 90\degree  & 180\degree & 270\degree  & ensemble \\
			
			\midrule
			\parbox[t]{1mm}{\multirow{2}{*}{\rotatebox[origin=c]{90}{}}}

			& IDEAL w/o splitting-head       & 82.1\% & 77.1\% & 78.6\% & 77.0\% & 85.5\% & 65.5\% & 61.1\% & 62.3\% & 61.4\% & \textbf{69.5}\% \\
			
			& IDEAL w/ splitting-head       & \textbf{84.0\%} & \textbf{80.2\%} & \textbf{82.8\%} & \textbf{80.5\%} & \textbf{87.1\%} & \textbf{65.9\%} & \textbf{62.2\%} & \textbf{62.9\%} & \textbf{62.0\%} & 68.7\% \\
			\bottomrule
		\end{tabular}
		\vspace{.2cm}
		\caption{Ablation experiments of dividing the last embedding layer on Cars-196 and CUB-200.}
		\label{splitting}
	\end{table}
	
	In Tab. \ref{splitting}, we explore how the effect of our method depends on the design of splitting the projection head. In this manner, the embedding dimension is split into $128$ for each domain for a fair comparison. 
	As illustrated in Tab. \ref{splitting}, without this design, Our IDEAL has a 2\% drop on Cars-196 and a slight drop on CUB-200 across all domains. This shows the benefits of sharing the low-level parameters only rather than the parameters of the whole network. Possibly, the multiple projection heads could aid the stable convergence because it would not suffer from conflicting gradient aggregated in high-level layers from different domains.

	
	
	
	\subsubsection{Combined with Exisiting DML Loss}
	
    Our IDEAL can be easily integrated into almost any existing DML loss to boost its performance. We evaluate our IDEAL combined with contrastive loss, triplet loss, and MS loss.  As shown in Tab. \ref{loss-table}, our IDEAL brings significant performance gain consistently on all datasets, over 3\%-5\% improvements. In particular, without bells and whistles, our IDEAL remarkably boost the performance: up to +\textbf{5.4} with contrastive, +\textbf{4.7} with triplet, and +\textbf{4.0} with MS loss. The results suggest that our IDEAL is orthogonal to the existing DML loss method. Both sophisticated and straightforward loss can get 3\%-5\% gain by our IDEAL.

	\begin{table*}[htb]
				\vspace{-0.3cm}
	\tablestyle{4.5pt}{1.0}
	\vspace{0em}
	\begin{tabular}{l|cccc|cccc|cccc}
		\multirow{2}[2]{*}{} & \multicolumn{4}{c|}{\multirow{2}[2]{*}{Cars-196}} & \multicolumn{4}{c|}{\multirow{2}[2]{*}{CUB-200}} & \multicolumn{4}{c}{\multirow{2}[2]{*}{SOP}} \\
		& \multicolumn{4}{c|}{}         & \multicolumn{4}{c|}{}                         & \multicolumn{4}{c}{}   \\
		Recall$@K$ (\%)    & 1     & 2    & 4    & 8     & 1     &   2    & 4    & 8     & 1    & 10    & 100   & 1000       \\ \shline
		Contrastive        & 75.8  & 84.4 & 90.3 & 94.2  & 61.8  & 73.1   & 81.9 & 89.2  & 59.1 & 75.8  & 87.3  & 95.4  \\
		\bf Contrastive w/ IDEAL & \bf 81.2  & \bf 87.8   & \bf  92.2  & \bf 95.4 & \bf 64.3   & \bf  75.3    & \bf  83.9    & \bf 89.6  &  \bf 63.2 & \bf 78.2 & \bf 88.1 & \bf 95.5 \\ \hline
		Triplet           & 75.5 & 83.1 & 88.4 & 92.7    & 60.7 & 72.0 & 81.6 & \bf 88.5         & 61.3 & 77.5 & 88.4 & 95.7 \\
		\bf Triplet w/ IDEAL & \bf 80.2  & \bf 87.0  & \bf 91.7 & \bf 94.4  & \bf 63.8   & \bf 73.5 & \bf  82.1 &  88.3 &  \bf 65.1  & \bf 80.2 & \bf 90.0  & \bf 96.5  \\ 
		\hline
		MS                & 83.1 & 89.9 & 93.9 & 96.6    & 64.5 & 76.1 & 85.1 & 90.8    & 78.2 & 90.3 & 96.0 & 97.3 \\
		\bf MS w/ IDEAL    & \bf 87.1 & \bf 92.6 & \bf 95.6 & \bf 97.4  & \bf 69.7 & \bf 79.5 & \bf 86.5 & \bf 91.3 &  \bf 81.1 & \bf 91.7 & \bf 96.6 & \bf 99.0   \\
	\end{tabular}%
	\vspace{.3em}
	
	\caption{Retrieval results of ours augmented (`w/ IDEAL')  methods compared with their respective baselines on three datasets.}
	\label{loss-table}%
	\vspace{-1.em}
	\end{table*}%

	\subsection{Comparison with State-of-the-Art} 

	\begin{table*}[htb]
		\small
		\centering
		\label{tab:cub-cars}
				\setlength\tabcolsep{4.5pt}
				\scalebox{0.78}[.8]{
		\begin{tabular}{lccccccccccc}
			\toprule
			\multirow{3}{*}{Methods} & \multicolumn{3}{c}{Cars-196~\cite{cars-196}} && \multicolumn{3}{c}{CUB-200~\cite{cub-200}} && \multicolumn{3}{c}{SOP~\cite{lifted-structured-loss}}\\
			\cmidrule{2-4} \cmidrule{6-8} \cmidrule{10-12}
			
			& R@1 & R@2 & R@4 &  & R@1 & R@2 & R@4  &  & R@1 &R@10& R@100 \\
			\midrule
			
			LiftedStruct$^{B}$~\cite{lifted-structured-loss}    &53.0 &65.7 &76.0  &&43.6 &56.6 &68.6  &&62.5 &80.8 &91.9 \\
			Clustering$^{B}$~\cite{struct-clustering}           &58.1 &70.6 &80.3  &&48.2 &61.4 &71.8  &&67.0 &83.7 &93.2\\
			ProxyNCA$^{B}$~\cite{movshovitz2017no}              &73.2 &82.4 &86.4  &&49.2 &61.9 &67.9  &&73.7 & -   & -  \\
			SmartMining$^{G}$~\cite{harwood2017smart}           &64.7 &76.2 &84.2  &&49.8 &62.3 &74.1  &&75.2 &87.5 &93.7\\
			HDC$^{G}$~\cite{yuan2017hard}                       &73.7 &83.2 &89.5  &&53.6 &65.7 &77.0  &&69.5 &84.4 &92.8\\
			Margin$^{R}$~\cite{wu2017sampling}                  &79.6 &86.5 &91.9  &&63.6 &74.4 &83.1  &&72.7 &86.2 &93.8\\
			HTL$^{B}$~\cite{HTL}                                &81.4 &88.0 &92.7  &&57.1 &68.8 &78.7  &&74.8 &88.3 &94.8\\
			ABIER$^{G}$~\cite{opitz2018deep}                    &82.0 &89.0 &93.2  &&57.5 &71.5 &79.8  &&74.2 &86.9 &94.0\\
			ABE$^{G}$~\cite{Kim_2018_ECCV_ABE}                  &85.2 &90.5 &94.0  &&60.6 &71.5 &79.8  &&76.3 &88.4 &94.8\\ 
			FastAP$^{R}$ ~\cite{FastAP}                         & -   & -  & -     && -   & -   & -    &&73.8 &88.0 &94.9 \\
			MIC$^{R}$  ~\cite{MIC}                              & -   & -  & -     && -   & -   & -    &&77.2 &89.4 &95.6 \\
			Multi-Simi$^{B}$~\cite{wang2019multi}               &84.1 &90.4 &94.0  &&65.7 &77.0 &86.3  &&78.2 &90.5 &96.0 \\ 
			SoftTriple$^{B}$~\cite{qian2019softtriple}          &84.5 &90.7 &94.5  &&65.4 &76.4 &84.5  &&78.6 &86.6 &91.8 \\
			CircleLoss$^{B}$ ~\cite{sun2020circle}              &83.4 &89.8 &94.1  &&66.7 &77.4 &86.2  &&78.3 &90.5 &96.1 \\
			
			XBM$^{B}$ ~\cite{wang2019cross}                     &82.0 &88.7 &93.1  &&65.8 &75.9 &84.0  &&79.5 &90.8 &96.1 \\
			
			\midrule \midrule
			\textbf{Multi-Simi$^{B}$ w/ IDEAL}   & \textbf{87.1} & \textbf{92.6} & \textbf{95.6}   && \textbf{69.5} & \textbf{79.5} & \textbf{86.5}    && \textbf{81.1} & \textbf{91.7} & \textbf{96.6} \\
			\bottomrule
		\end{tabular}}
		\caption{Comparison with state of the art on Cars196, CUB200 and SOP. Superscripts denote backbone networks by abbreviations, G: GoogleNet,B: BN-Inception, R: ResNet50.}
		\label{sota}
	\end{table*}
	 
	In this section, we compare our IDEAL with the SOTA methods on three established benchmarks. Tab. \ref{sota} summarizes our results. Our IDEAL with MS loss performs better than all competitors by a large margin. On the large dataset, SOP, our method surpass the current SOTA methods: XBM by $79.5 \% \rightarrow 81.1 \%$. Our IDEAL with BN-Inception even outperforms existing methods with ResNet50 considerably, such as FastAP by 73.8\%$\rightarrow$81.1\%. On Cars and CUB, our IDEAL also achieves the best results of 87.1 \% and 69.5 \%. Furthermore, our work improves all of the previous ensemble methods largely. For example, compared with ABE, improving performance by 2\%, 9\% and 5\% on Cars, CUB and SOP, respectively.

	\section{Conclusion}
	
	In this paper, we first defined a novel augmentation called \emph{Domain Augmentation} for general DML methods. Then we proposed our IDEAL mechanism, which can simultaneously learn multiple independent embedding spaces for multiple domains generated by predefined data transformations. We validate that our IDEAL is well suited for the general DML framework and achieve surprising performance in a variety of datasets. Notably, we do not explore orthogonal factors (such as alternative augmentation, \eg, rotations by multiples of 45 degrees) that may further improve performance. In the future, to improve its computational efficiency during testing, we will explore other transformations of domain augmentation and extensions of our training scheme (\eg. distillation techniques).

	\section*{Broader Impact}
	
	This work would generate crucial insights into the different training mechanisms employing varied data transformations beyond data augmentation, thus providing a basic scientific knowledge necessary for representation learning. Moreover, the results of this research would significantly advance understanding to the low-level but fundamental data transformation in the general area of computer vision. This knowledge is important for the development of sub-field in person re-identification, face recognition and image retrieval.
	
	

	\begin{ack}
		Use unnumbered first level headings for the acknowledgments. All acknowledgments
		go at the end of the paper before the list of references. Moreover, you are required to declare 
		funding (financial activities supporting the submitted work) and competing interests (related financial activities outside the submitted work). 
		More information about this disclosure can be found at: \url{https://neurips.cc/Conferences/2020/PaperInformation/FundingDisclosure}.

		Do {\bf not} include this section in the anonymized submission, only in the final paper. You can use the \texttt{ack} environment provided in the style file to autmoatically hide this section in the anonymized submission.
	\end{ack}
	
	
	
	
	\bibliographystyle{plainnat}
	\bibliography{egbib}

	
	
	
\end{document}